# Informing Computer Vision with Optical Illusions

Nasim Nematzadeh
College of Science and Engineering
Flinders University
Adelaide, Australia
nasim.nematzadeh@flinders.edu.au

David M. W. Powers
College of Science and Engineering
Flinders University
Adelaide, Australia
david.powers@flinders.edu.au

Trent Lewis
College of Science and Engineering
Flinders University
Adelaide, Australia
trent.lewis@flinders.edu.au

*Abstract*— Illusions are fascinating and immediately catch people's attention and interest, but they are also valuable in terms of giving us insights into human cognition and perception. A good theory of human perception should be able to explain the illusion, and a correct theory will actually give quantifiable results. We investigate here the efficiency of a computational filtering model utilised for modelling the lateral inhibition of retinal ganglion cells and their responses to a range of Geometric Illusions using isotropic Differences of Gaussian filters. This study explores the way in which illusions have been explained and shows how a simple standard model of vision based on classical receptive fields can predict the existence of these illusions as well as the degree of effect. A fundamental contribution of this work is to link bottom-up processes to higher level perception and cognition consistent with Marr's theory of vision and edge map representation.

## I. INTRODUCTION

Optical illusions highlight the sensitivities of human visual processing and studying these leads to insights about the perception that can potentially illuminate the ways we can implement bioplausible approaches in computer vision. Geometrical illusions [1] are a subclass of illusions in which orientations and angles are distorted and misperceived. Tile Illusions are Geometric Illusions with second-order tilt effects which arise from the contrast of background and tilt cues such as in the Café Wall illusion. A quantified model for detecting the illusory tilts in the Café Wall illusion has been reported [2, 3] and we are going to generalize this approach for more complex Tile Illusions such as Complex Bulge pattern which was introduced in [4, 5] and more generally to a range of Geometric Illusions that will be explored as the primary new contribution in this work.

Visual processing starts with the sensations of the retinal receptive fields (RFs) by the incoming light into the eyes. Retinal ganglion cells (RGCs) are the retinal output neurons that convert synaptic input from the inner plexiform layer (IPL) and carry the visual signal to the brain. The diversity of RGC types and the size dependence of each specific type to the eccentricity (the distance from the fovea) are physiological evidence [6] for multiscale encoding of the visual scene in the retina. Consequently, low-level computational models of retinal vision have been proposed based on the simultaneous sampling of the visual scene at multiple scales [7]. Even given the increasingly detailed biological characterization of both retinal and cortical cells over the last half a century (1960s-2010s), there remains considerable uncertainty, and even some controversy, as to the nature and extent of the encoding of visual information by the retina, and conversely of the subsequent processing and decoding in the cortex [6, 8].

We explore the response of a simple bioplausible model of low-level vision on Geometric/Tile Illusions, reproducing the misperception of their geometry, that we reported for the Café Wall and some Tile Illusions [2, 5] and here will report on a range of Geometric illusions. The model has until now not been verified to generalize to these other illusions, and this is what we show in this paper.

Although the misperception of orientation in Tilt Illusions in general, may suggest physiological explanations involving orientation selective cells in the cortex (such as in [9]), our work provide evidence for a theory that the emergence of tilt in these patterns is initiated before reaching orientation-selective cells as a result of known retinal/cortical simple cell encoding mechanisms. Our experimental results suggest that Differences of Gaussian (DoG) filtering at multiple scales has a significant role in explaining the induced tilt in Tile Illusions in general and has the potential to reveal some of the illusory cues we perceive in some Geometrical illusions in particular.

## II. THE MODEL

### A. The DoG Model Specifications

It is shown by numerous physiological studies that what is sent to the cortex is a multiscale encoding of the visual scene as a result of the diverse range of the receptive field types and sizes inside the retina [6, 8, 10]. We can model the activations of the retinal/cortical simple cells by using Differences and/or Laplacian of Gaussians (DoG, LoG) introduced by vision pioneers such as Rodieck and Stone [11] and Enroth-Cugell & Robson [12]. For modelling the RFs, it is also shown that the DoG filtering is a good approximation of LoG if the scale ratio of surround to center Gaussian is close to 1.6 [13, 14].

Therefore, the RGC responses can be modeled by a DoG transformation that creates an edge map representation at multiple scales for a given pattern used in our studies [2, 3]. The parameters of the model are the scale ratio of the surround to the center Gaussian in the DoG filter that referred to as *Surround ratio* ($s$). In respect to the filter size, we need to make sure that the DoG is only applied within a window in which the value of both Gaussians are insignificant outside the window and to control this we use another parameter called *Window* *ratio* (*h*). This parameter determines how much of each Gaussian is included inside the filter. $s = 2.0$ is typically used in our model [2, 3]. The ratio of 1:1.6 – 2.0 indicates the size of center: surround Gaussians is a typical range for modeling simple cells. (Marr and Hildreth [13] used 1:1.6 for modeling retinal GCs in general, and this ratio is used by Earle and Maskell [14] for DoG modeling specifically to explain the Café Wall illusion. Here instead of a Surround Ratio of $s = 2.0$ used in the previous reports [2, 3], we have used $s = 1.6$ [13].

Our model creates a raw-primal representation for Tile Illusions (patterns investigated in this research) and for many other types of visual data, referred to as the '*edge map*' represented at multiple scales. The sigma of the center Gaussian ($\sigma_c$) is the central parameter of the model and its optimal value is dependent on the dimensions of the pattern's elements. For instance, to extract the tilted line segments along the mortar lines in the Café Wall illusion, $\sigma_c$ should be similar in size to the mortar thickness [3].

We further investigate the edge maps of four variations of Geometrical illusions which consist of the Hermann Grid, Zöllner, Spiral Café Wall, and Complex Bulge illusions in Section B, to evaluate the efficiency of our low-level filtering approach for modelling the induced tilt effects we perceive in them. Then we explain more on detecting tilts from the edge maps of two illusions of the Spiral Café Wall and Complex Bulge patterns in order to quantify the tilt effects in them in Section III. Finally in Section IV we investigate the edge maps for two samples of scenery images and explain the details about the Marr's vision theory and his speculation of the primal sketch representation.

*B. DoG edge maps of well-known illusory patterns*

The Classical Receptive Field (CRF) is the area in which a visual stimulus evokes a charge in the firing activity of a cell, which can explain perceptual effects in illusions such as the Hermann Grid and Mach Bands [15-18]. The model output as the edge maps for four Illusions of the Hermann Grid, Zöllner, Spiral Café Wall, and Complex Bulge patterns have been provided in Figs 1 and 2 based on a constant value of $s = 1.6$ and $h = 8$ as the *Surround and Window ratios* in the model. Although we investigated the DoG edge maps of these patterns at fifteen different scales in our experiments ($\sigma_c = 1$ to 15 with incremental steps of 1), we presented the edge maps at every second scale (8 scales in the figures) for the sake of saving the space.

*a) Hermann Grid*

The classical explanation of the appearance of Grey spots in the intersections of horizontal and vertical Black bars in the Hermann Grid illusion [18, 19] is that when the retinal ganglion cells (RGCs), positioned at the crossings (intersections), the effect of inhibitory surround would be four, but when the RGC is looking at a street (Bars but not their intersections) it gets only two inhibitory patches, so it will have a higher spike rate compared to the ones at the crossings. This was measured by Baumgartner in 1960 [20]. So inhibitory response occurs as a result of increasing the white surround in the pattern. Spillmann used the Hermann Grid stimulus to estimate the size of visual receptive fields in man and tested a modified version of that with 15 shades of grey for the bars, viewed against different uniform backgrounds to measure the contrast sensitivity of the subjects in perceiving the illusion [18]. The pattern may have the opposite arrangement compared to the investigated pattern with White bars on top of a Black background.

Fig. 1 (Left) shows the DoG edge maps in jetwhite color map [21] and in the binary form at eight different scales from $\sigma_c = 1$ to 15 with incremental steps of 2, for the Hermann Grid illusion. The pattern has a size of 512×512px with the bar widths of 16px. There is no tilt effect in this pattern, but the explanation for the appearance of flashing Grey spots (dots) in the intersections of Black bars which has been given in the literature is the result of lateral inhibition (LI) and ON and OFF center-surround activity of the retinal GCs. The Grey spots appear in the peripheral view of the pattern and disappear at the focal view resulting in flashing Grey spots [18, 22].

Our explanation considers both local and global views of the pattern at fine to coarse scale DoG edge maps, which simulate foveal to peripheral ganglion cells (GCs) in the retina. The model's DoG edge map for the pattern nicely illustrates this explanation for the appearance of Grey spots in this pattern. Considering the jetwhite representation of the edge map, at fine scales ($\sigma_c = 1$ to 3), we see light-Blue spots appear at the intersections of bars. By increasing the DoG scale ($\sigma_c$), a transient state emerges (from light-Blue to dark-Blue spots in the intersection points of bars on the edge map). It is worth mentioning that the brightness level of these spots (dots) relates to the DoG convolution and its averaging process, as well as the position of DoG filter and its size (scale) for generating the edge map representation. The color bar on the right side of the jetwhite representations of the edge maps in Figs 1 and 2 facilitates navigation on the DoG convolved outputs (for example it shows that both light-Blue and dark-Blue are negative values, with dark-Blue being more negative). In the binary form, the appearance of spots is not as clear as in the jetwhite representation, but we see dark Grey dots at fine scales and light-Grey dots at coarse scales in the edge map simulating larger receptive fields (RFs).

Due to the simultaneous sampling of RGCs and multiscale retinal encoding of the pattern, the visual appearance of spots is not persistent and seems to flash while shifting our gaze on the pattern. At focal view to the intersections, due to very fine scale RGCs in the fovea, we get a sharp fine scale response with no spots visible in the intersections of the bars, similar to $\sigma_c = 1$ in Fig. 1 (Left) in our simulations. It seems that the final perception of the pattern is affected by our local to the global view of the pattern, highlighting the appearance of flashing Grey spots in the intersections of Black bars.

*a) Zöllner illusion*

The Zöllner illusion [23] consists of a series of parallel Black lines in a diagonal orientation, intersected by a pattern of short inducing line segments, alternating between horizontal and vertical along the long parallel Black lines. These short inducing lines create an illusory percept of the long Black lines as not being parallel. The angle of short line segments to the longer lines results in this impression of converging and diverging illusory effects of long parallel lines. It has been claimed in some literature that the maximum effect is when the acute angle of inducing lines and long lines is between 15° to 30° [24]. This Geometric Illusion is regarded as one of the Café Wall illusions [25-27].

The edge maps of this pattern at multiple scales are presented in Fig. 1 (Right) in the jetwhite color map [21] and in the binary form with DoG scales similar to the Hermann Grid illusion on the left. The Zöllner pattern investigated has a size of 540×540px. As the DoG edge map reveals, at fine to medium scales ($\sigma_c$ = 1 to 7), the fine details of short inducing line segments are detected, preserving two opposite orientations of either vertical or horizontal direction along each long parallel lines (diagonal lines). As the scale increases, around scale 9, the orientation of the small line segments starts to fade and after scale 11 they are completely lost in the edge map. What we see are nearly similar wiggling lines (zigzag lines on the long diagonal parallel lines), which are the integration of DoG outputs of the short inducing line segments, joined together at these scales of the DoGs along the long diagonal parallel lines. By increasing the DoG scale to scale 15, what the DoG edge map reveals is just long slanted lines in a diagonal orientation, with no cues of intersected short inducing lines on top of them. We believe that the perception of converging and diverging of long parallel slanted lines in the pattern (with the diagonal arrangement) is the result of multiscale encoding of the retinal cells and incompatible grouping of pattern elements that happen simultaneously at different scales, which contribute to the illusory tilt perception in the pattern.

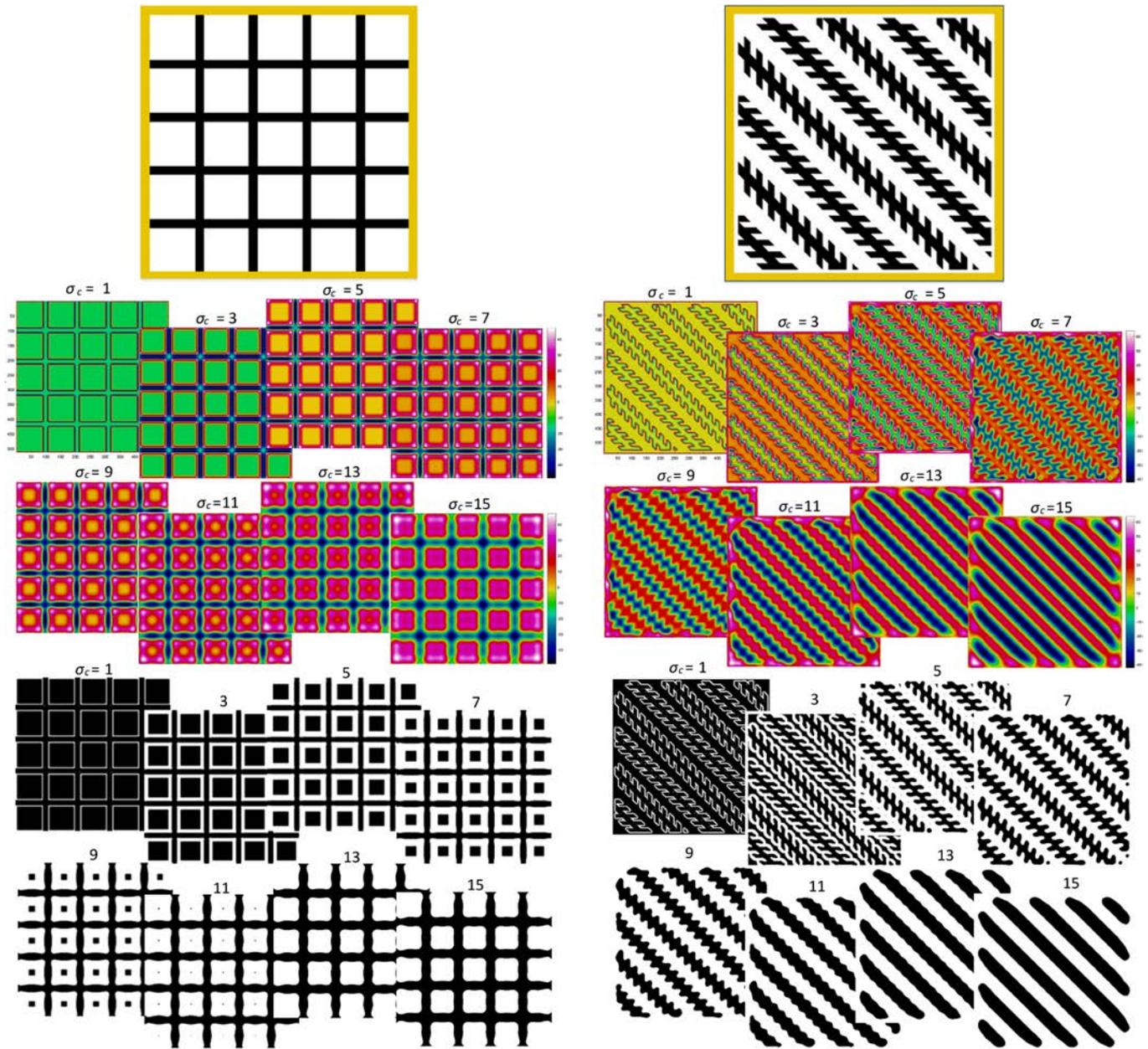

**Fig. 1. (Left)** - Top: Hermann Grid illusion, framed in Yellow on top, Middle and Bottom: The DoG edge maps of the pattern at eight different scales from left to right and top to bottom ($\sigma_c$ = 1 to 15 with incremental steps of 2), presented first in the jetwhite color map (in the middle), and then in the binary form (at the bottom). The investigated pattern has a size of 512×512px, with bar widths of 16px. **(Right)** - Zöllner illusion on top followed by the DoG edge maps of the pattern at eight different scales, in the jetwhite color map (in the middle), and in the binary form (at the bottom). The investigated pattern has a size of 540×540px. $s$ = 1.6, $h$ = 8 (*Surround* and *Window* ratios respectively) are constant in the DoG model here.

*b) Spiral Café Wall*

In the Spiral Café Wall illusion [28] the rows of Café Wall tiles are arranged in a circular way with mortar lines in between (like rings of tiles), and we get an impression of their arrangement as being in a spiral organization that is an illusory percept of the pattern. Fig. 2 (Left) shows the edge maps for the pattern with the size of 875×875px ($\sigma_c$ = 1 to 15 with incremental steps of 2). The pattern is among the complex Tile Illusion patterns compared to the Café Wall illusion [5]. Due to the circular design of the pattern, the size of tiles and mortar lines has been increased from the center to the surround region of the pattern and encoding of tiles as well as mortar lines are different from the original Café Wall pattern which consists of constant tile size and mortar size. The appearance of a complete tile on the edge map is dependent on the scale of the edge map relative to each individual tile size in a ring of tiles in this pattern. This is the same for the mortar lines as well.

What we see at fine scales of the edge map is the extraction of fine details from mortar lines to tiles edges in the pattern. The tiles are connected through mortar lines at very fine scales ($\sigma_c \leq 3$). As the scale increases, we see some blending of color in the DoG output and that the mortar cues start to disconnect. As a result, the grouping of tiles by the mortar lines disappears and a different grouping of tiles starts to be revealed at scale 5 for the central tiles, and near scale 7 for the peripheral ones close to the outer border of the pattern. The new grouping of tiles generates curved lines, from the center of the pattern to the periphery, which is thin at the center and thick at the other end. Similar groups appear around the whole pattern, moving from the center to the outer region with a slight circular rotation of a similar curved line at mid to coarse scales in the edge map. Another thing worth mentioning is the size extension of the central hole in the DoG edge map as the scale increases. All of these cues can contribute to the perception of the Spiral Café Wall rather than the circular arrangement of the Café Wall tiles in the pattern.

*c) Complex Bulge pattern*

The Complex Bulge pattern [29] consists of a simple checkerboard background and some superimposed White/Black dots on Black/White tiles, arranged in the center of the checkerboard, giving the impression of a central bulge in the pattern. The superimposed dots on their backgrounds give some impression of foreground-background percept. Different positions of dots on the textured background result in some tilt, bow or wave perceptions along the edges as well as expansion and contractions on checkers corners as noted in [30].

The edge maps of the Complex Bulge pattern with the resolution of 574×572px are presented in Fig. 2 (Right) in the jetwhite color map and in the binary form at eight scales ($\sigma_c$ = 1 to 15 with incremental steps of 2). At the finest scale ($\sigma_c$ = 1), the DoG output reveals the fine details of the pattern including the edges of tiles and the superimposed dots. What we see at fine scales is a grouping of tiles with superimposed dots in a circular arrangement around the center. The impression of central bulge in the edge map lasts till nearly scale 5, and at this scale we see a transient state from the central bulge effect as the result of the grouping of tiles with superimposed dots to a different grouping of tiles in an X shape organization of identically colored tiles in the pattern. This grouping of tile elements persists from mid to coarse scales when the cues of fine-scale superimposed dots have been disappeared in the edge map.

Again, what we believe as the explanation of the illusory percept of the central bulge in the pattern is the simultaneous sampling of the pattern elements at multiple scales in the retinal encoding of the pattern, which results in two incompatible groupings of pattern elements which contribute to the illusory perception of a central bulge in the pattern. Based on the relative size of superimposed dots and the tiles of the checkerboard, we see the impression of central bulge very clearly for the given pattern. It is obvious that decreasing the size of superimposed dots results in less persistency of dot cues at fine scales of the DoG edge map, and a weaker illusory bulge effect at the center [30].

### III. DETECTING TILTS IN THE EDGE MAPS

Our model provides new insights into physiological models [6, 8, 31] as well as supporting Marr's theory of low-level vision [13, 32]. We briefly illustrate in this section how we further analyse the detected tilted lines in the DoG edge maps of the investigated patterns at multiple scales using the Hough analysis as the second stage processing [33] on these illusions. The model's early stage output is investigated to quantify the degree of tilt using the Hough Transform [34] in place of the later higher-order cortical processing. Mean tilt and standard deviation of the detected tilted line segments are calculated for every scale of the edge map (for example in [33]), providing quantified predictions for these experiments with human subjects.

Two outputs of this stage are presented in Fig. 3, as detected houghlines shown in Green, displayed on the binary edge maps at multiple scales. The parameters of the model are $s$ = 1.6 and $h$ = 8 here. The DoG scales ($\sigma_c$) are presented at the top of each row of the edge maps indicating the scales from left to right for the DoG-filtered outputs in the figure. On the top of Fig. 3, the detected houghlines are shown in Green, displayed on the DoG edge map of the Spiral Café Wall illusion at twelve different scales (from $\sigma_c$ = 0.5 to 6.0 with incremental steps of 0.5). The hough parameters for detecting tilt angles are provided in the figure caption. At fine scales ($\sigma_c$ = 0.5 to 1.5), we see small detected line segments connecting the tiles' edges with the mortar lines in a circular manner around the center and since the edges are dense at the central part of the pattern, the hough algorithm detects other lines with another direction from the center to the outer region, concentrated in the central part of the pattern at the finest scale ($\sigma_c$ = 0.5) which extends to the surround region as the scale increases. At scales 1.5 and 2.0, we see similar distributions of these two different lines with a circular organization around the center (connecting tiles with mortar lines) as well as small line segments whose integrations result in some kind of curved lines from the center to the periphery of the pattern, connecting tiles together in the regions where the mortar cues are disappeared in the edge map.

At medium scales ($\sigma_c$ = 2.5 to 4.0)-second row of the edge map (with modified hough parameters), we can see these lines more clearly, showing how the majority of houghlines are detected along the center to the periphery with their integration resulting in construction of curved lines from the center to the surround region of the pattern (concentrated on the central regions and up to the middle of the pattern) where the mortar cues start to fade. In the outer region of the pattern, when the mortar cues still exist in the edge map, the houghlines are detected connecting tiles with the mortar lines in nearly circular arrangements. Also, we can see the increasing size of the central hole in the edge map of the pattern by increasing the DoG scale. At coarse scales ($\sigma_c$ = 5.0 to 6.0), the integrated curved lines are extended to the border of the pattern. We have shown here that the result of the detected houghlines matches the tilt cues as they appear in the edge map of the pattern at multiple scales. This is exactly what we had aimed to find an explanation for the tilt effect in the pattern, highlighting different groupings of pattern elements at different scales of the edge map, as we can observe across multiple scales.

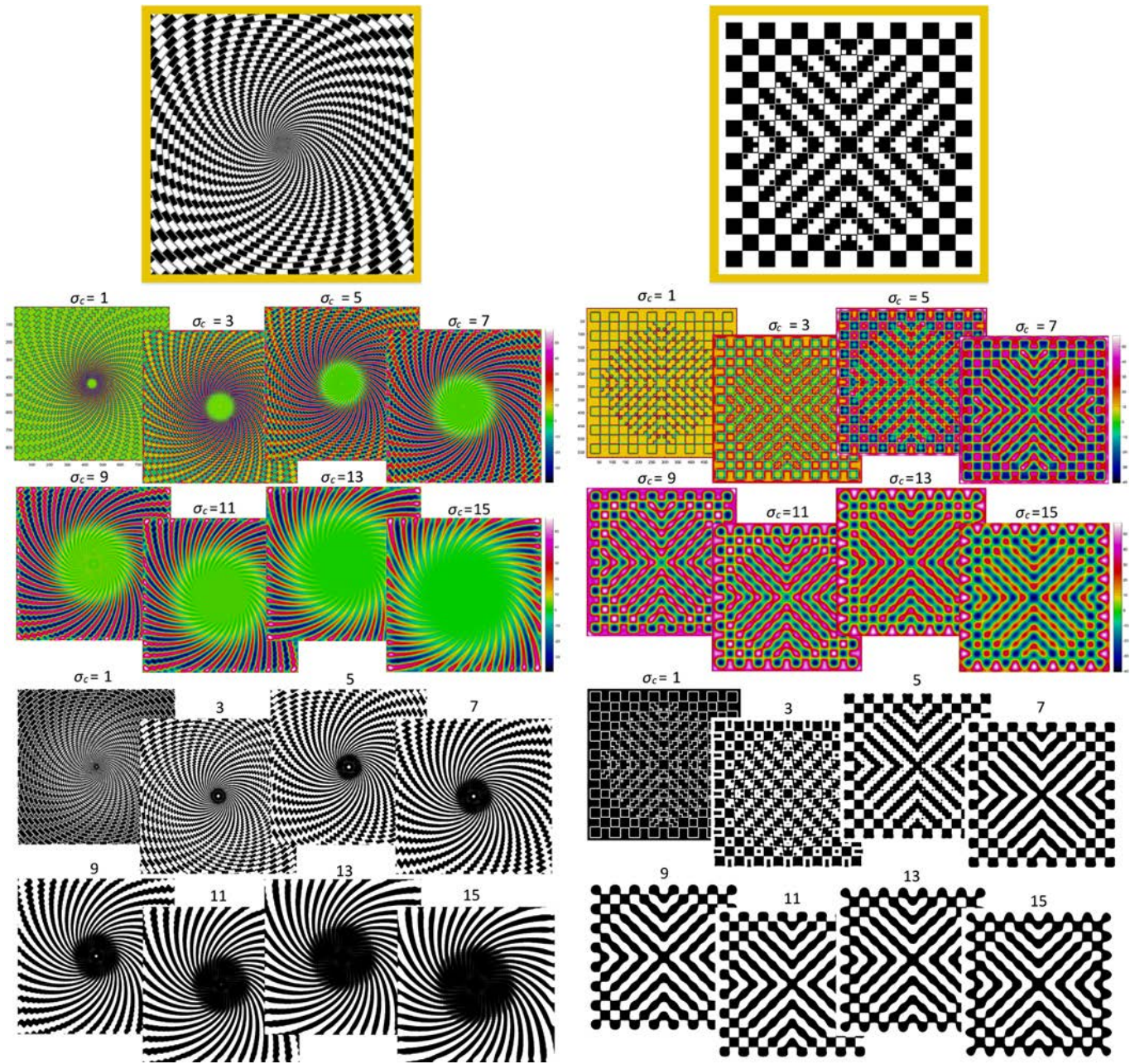

**Fig. 2. (Left)** - Top: Spiral Café Wall illusion, framed in Yellow on top, Middle and Bottom: The DoG edge maps of the pattern at eight different scales from left to right and top to bottom ($\sigma_c$ = 1 to 15 with incremental steps of 2), presented first in the jetwhite color map (in the middle), and then in the binary form (at the bottom). The investigated pattern has a size of 875×875px. **(Right)** - Complex Bulge pattern on top followed by the DoG edge maps of the pattern at eight different scales, in the jetwhite color map (in the middle), and in the binary form (at the bottom). The investigated pattern has a size of 574×572px. $s$ = 1.6, $h$ = 8 (*Surround* and *Window ratios* respectively).

The bottom of Fig. 3, shows the detected houghlines for the Complex Bulge pattern in Green, displayed on the edge map at eight scales from $\sigma_c$ = 1.0 to $\sigma_c$ = 4.5, with incremental steps of 0.5. The hough parameters are provided in the figure caption. As the detected houghlines show in the figure, at fine to medium scales-top row, the detection of slanted line segments on the edge map starts at scale 1.5 with their integration resulting in a central bulge from the center outwards. This is getting clearer at scale 2.0, with the detected tilted lines around the center with an impression of bulge and then the detection of the tiles' edges out of the central region with no superimposed dots. As the scale increases ($\sigma_c$ = 3 to 4.5-second row), hough detects other line segments in nearly diagonal orientations (positive and negative), originating at the center whose integration results in an X shape grouping of these lines. Again our proposed hough analysis correctly detects the tilt cues as they appear in the DoG edge map of this pattern at multiple scales.

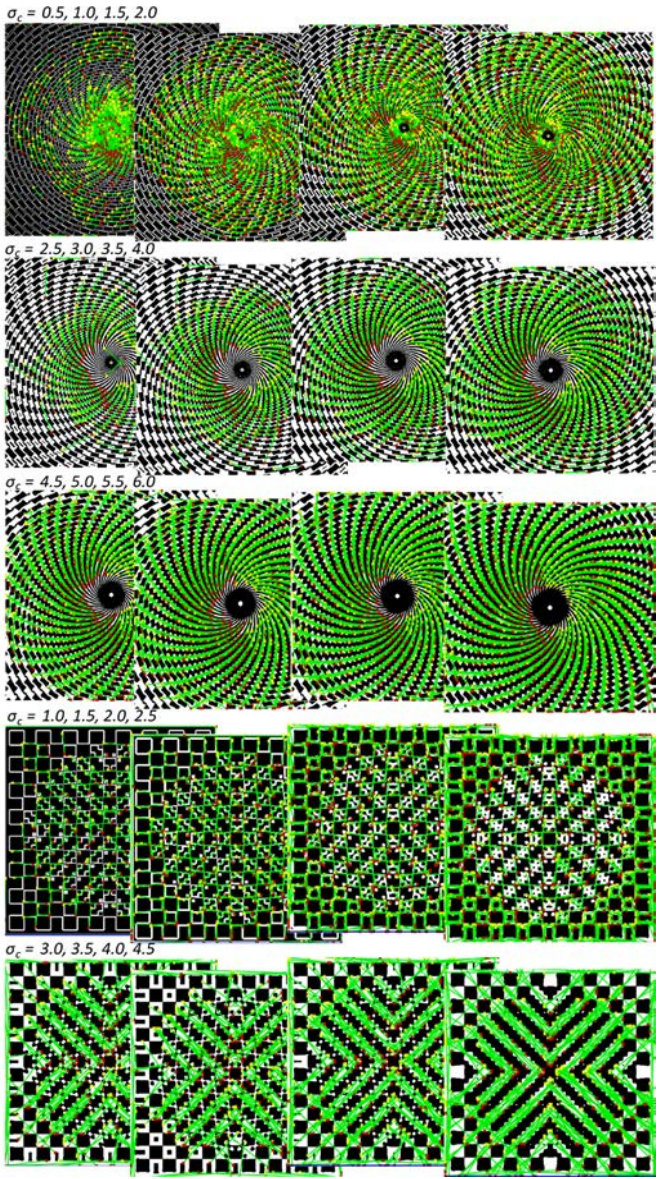

**Fig. 3.** Edge/tilt detection for two stimuli from their edge maps at multiple scales. **(Top)** Tilt detection for the Spiral Café Wall illusion ($\sigma_c$ = 0.5 to 6.0 with incremental steps of 0.5) with hough parameters of FillGap=5, MinLenght = 50 (1st row) and FillGap = 5, MinLenght = 100 (2nd and 3rd Row). **(Bottom)** Tilt detection for the Complex Bulge pattern ($\sigma_c$ = 1 to 4.5 with incremental steps of 0.5) with hough parameters of FillGap = 5, MinLenght = 50 (1st row) and FillGap = 10, MinLenght = 100 (2nd row). Crosses mark start (yellow) and end (red). Other parameters of the model are $s$ = 1.6, and $h$ = 8.

IV. RELATION TO MARR'S MODEL

The description of simple cells in Hubel and Wiesel's [35] work described as the bar- or edge-shaped receptive fields led to a view of the population of feature detectors of edges and bars of various widths and orientations in the cortex explained by Barlow [36]. On the other hand, Campbell and Robson's [37] experiments processed images in parallel with a number of independent orientation and spatial frequency-tuned channels, highlighted a different perspective in which the visual cortex performs a kind of spatial Fourier analysis [13]. Considering these two views, Marr and Hildreth note that none of these approaches can provide any direct information about the goals of the early analysis of an image [13].

Marr and Hildreth emphasized that "the purpose of early visual processing is to construct a primitive but rich description of the image that is to be used to determine the reflectance and illumination of the visible surfaces and their orientation and distance relative to the viewer" [13- pp.188]. Marr and his colleagues introduced the first primitive descriptor of the image as the primal sketch [38] that is formed in two parts: (1) Intensity changes using edge segments, bars, blobs and terminations referred to as raw primal sketch. (2) Geometrical relations, with more abstract tokens by selecting, grouping, and summarizing the raw primitives in various ways. These two results in a hierarchy of descriptors, covering a range of scales, referred to as full primal sketch of an image.

In Marr's theory of edge detection [13], in order to detect the intensity changes over a wide range of scales for natural images, they used the second derivative of Gaussian filter (LoG) which does not need to be orientation-dependent. The intensity changes in each of the channels are then presented by zero-crossing segments which are oriented primitives. Marr and Hildreth further noted that the zero crossing segments from different channels are not independent and in the image description they should combine.

They also demonstrated how the aggregate response of a group of ON- and OFF- receptive fields can produce the directional selectivity properties in the response. They noted: "If P presents an ON-center Geniculate X-cell receptive field, and Q, an OFF-center one, then if both are active, a zero-crossing Z is the Laplacian passing between them. If they are connected to a logical AND gate, as shown, then the gate will 'detect' the presence of the zero-crossing. If several are arranged in tandem, as in (b) and also connected by logical ANDs, the resulting operation detects an oriented zero-crossing segment within the orientation bounds given roughly by the dotted lines. This gives our most primitive model for simple cells. Ideally one would like gates such that there is a response only if all (P, Q) inputs are active, and the magnitude of the response then varies with their sum" [13-Fig. 9 Caption].

In comparison to Marr's and Hildreth's edge descriptor based on finding the zero-crossing segments of the LoG filtered output, we have implemented a DoG approximation to it with a specified scale ratio of the Gaussians as described by them ($\sigma_i/\sigma_e$ = 1.6 [13]; where i is inhibitory/surround, and e is excitatory/center). The binary edge map in our simulations is quite similar to the LoG output of Marr's and Hildreth's model prior to finding the zero-crossings. Fig. 4 shows the DoG edge map of the model for a sample natural image at only four selected scales from an edge map with 12 scales ($\sigma_c$ = 3 to 12 with incremental steps of 3 instead of 1 for the range of $\sigma_c$ = 1 to 12). The range of the DoG scales is selected empirically in this experiment in a way to capture nearly a full range of fine to medium scale features from the image. The simulation results show that the model output is not very sensitive to precise parameter setting. To extract an optimized and rich edge descriptor, we can use Logan's theorem (1977) as described by Marr and Ullman [39] to find the zero crossing of one-octave bandpass signals [13].

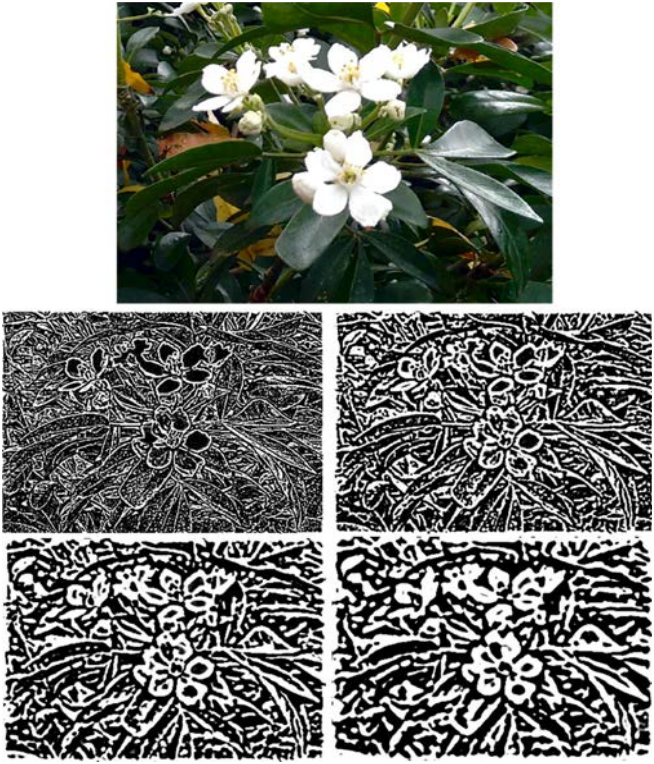

**Fig. 4.** The DoG edge map presented at four different scales ($\sigma_c$ = 3 to 12 with incremental steps of 3) for a sample of natural image. The image size is 2250×1570px, and the constant parameters of the model are $s$ = 1.6, $h$ = 2.

We want to note that Marr's theory of edge detection based on modelling the activations of simple cells can not only define a primitive rich descriptor for the visual scenery in our field of view, but also it can reveal some of the illusion effects we perceive in some Geometrical illusions as the results of simple cells processing. The edge map representation in our simulations is consistent with Marr's theory of vision and his speculation of the simple cells encoding to include directional selectivity properties in the edge map. The simulation results show how a group activation of simple cells can encode tilt effects in Tile Illusions, which commonly assumed as the result of the orientation selectivity properties of more complex cells, and also some illusion cues we see in Geometric Illusions.

V. SUMMARY

The current models for Geometrical illusions are quite complicated and more research is needed to improve models of vision while directing them towards less sophisticated and more bioplausible detection of visual cues and clues. We believe that further exploration of the role of simple Gaussian-like models [13, 40, 41] in low level retinal processing, and Gaussian kernels in early stage DNNs [42], and its prediction of loss of perceptual illusion will lead to more accurate computer vision techniques and models and can potentially steer computer vision towards or away from the features that humans detect. These effects can, in turn, be expected to contribute to higher-level models of depth and motion processing and generalized to computer understanding of natural images.

As Marr and Hildreth described how a group activations of ON- and OFF- receptive fields can encode the directional selectivity in the aggregate response of these cells, we have shown this for the illusory tilts we perceive in Tile Illusions in general and some illusion effects in other Geometric Illusions such as the Hermann Grid and Zöllner illusion in particular. Based on the experimental results we note that more sophisticated models of non-classical receptive fields (nCRFs) with anisotropic filters such as [9] are not essential to reveal the illusion effects in these patterns, although we should note that for the final perception of these illusory tilts and the integration of the local tilt cues, higher level cortical processing by more complex cells is required. We also showed that in all these illusions investigated, two or more incompatible groupings of pattern elements arise simultaneously in their edge maps as a result of our local and peripheral views of these stimuli. That is a major factor which contributes to the induced tilt in these illusions.

We suggest that for complex Tile Illusions with inducing tilt effects and a broader range of Geometrical illusions similar to the investigated patterns that contain a diverse range of tilt and brightness/contrast cues, some kind of fusion of multiscale (local and global) representations of the input pattern will be required, considering the focus point (change of illusory effect with saccade) as well as a holistic view of the pattern for a complete explanation. A psychophysical assessment of the model predictions will also help the design of an analytical model to search for different visual clues in natural or illusion patterns similar to our visual processing.

REFERENCES

[1] J. Ninio, "Geometrical illusions are not always where you think they are: a review of some classical and less classical illusions, and ways to describe them," *Front Hum Neurosci,* vol. 8, p. 856, 2014.

[2] N. Nematzadeh and D. M. W. Powers, "A Predictive Account of Café Wall Illusions Using a Quantitative Model," *submitted; arXiv preprint:1705.06846,* 2017.

[3] N. Nematzadeh and D. M. W. Powers, "The Cafe Wall Illusion: Local and Global Perception from multiple scale to multiscale," *Journal of Applied Computational Intelligence and Soft Computing: Special issue of Imaging, Vision, and Pattern Recognition,* 2017.

[4] N. Nematzadeh, D. M. W. Powers, and T. W. Lewis, "A Neurophysiological Model for Geometric Visual Illusions," *NeuroEng 2016: 9th Australasian Workshop on Neuro-Engineering and Computational Neuroscience. ,* 2016.

[5] N. Nematzadeh, D. M. W. Powers, and T. W. Lewis, "Bioplausible multiscale filtering in retino-cortical processing as a mechanism in perceptual grouping," *Brain Informatics, DOI 10.1007/s40708-017-0072-8,* 2017.

[6] G. D. Field and E. J. Chichilnisky, "Information processing in the primate retina: circuitry and coding," *Annu Rev Neurosci,* vol. 30, pp. 1-30, 2007.

[7] T. Lindeberg and L. Florack, "Foveal scale-space and the linear increase of receptive feld size as a function of eccentricity," *KTH Royal Institute of Technology,* 1994.

[8] T. Gollisch and M. Meister, "Eye smarter than scientists believed: neural computations in circuits of the retina," *Neuron,* vol. 65.2 pp. 150-164, 2010.

[9] D. M. Todorovic, "A Computational Account of a Class of Orientation Illusions," *MODVIS2017 - Computational and Mathematical Models in Vision,* 2017.

[10] J. L. Gauthier *et al.*, "Receptive fields in primate retina are coordinated to sample visual space more uniformly," *PLoS Biol,* vol. 7, no. 4, p. e1000063, Apr 07 2009.

[11] R. W. Rodieck, Stone, J., "Analysis of Receptive Fields of Cat Retinal Ganlion Cells," *Journal of Neurophysiology, 28(5), 833-849,* 1964.

[12] C. Enroth-Cugell and J. G. Robson, "The contrast sensitivity of retinal ganglion cells of the cat," *J Physiol,* vol. 187, no. 3, pp. 517-52, Dec 1966.

[13] D. Marr and E. Hildreth, "Theory of edge detection," *Proc R Soc Lond B Biol Sci,* vol. 207, no. 1167, pp. 187-217, Feb 29 1980.

[14] D. Earle and S. Maskell, "Twisted Cord and reversall of Cafe Wall Illusion," *Perception, 22(4), 383-390,* 1993.

[15] J. M. H. du Buff, "Ramp edges, mach bands, and the functional significance of the simple cell assembly," *Biological Cybernetics,* vol. 70(5), pp. 449–461, 1994.

[16] L. Pessoa, "Mach Bands How Many Models are Possible-Recent Experimental Findings and Modeling attemps," *Vision Research, 36(19), 3205-3227,* 1996.

[17] F. Ratliff, "Mach bands: quantitative studies on neural networks," *Holden-Day, San Francisco London Amsterdam,* 1965.

[18] L. Spillmann and J. Levine, "Contrast enhancement in a Hermann grid with variable figure-ground ratio," *Exp Brain Res,* vol. 13, no. 5, pp. 547-59, Nov 30 1971.

[19] L. Hermann, "Eine erscheinung simultanen contrastes," *Pflügers Archiv European Journal of Physiology,* vol. 3(1), pp. 13-15, 1870.

[20] G. Baumgartner, "Indirekte grössenbestimmung der rezeptiven felder der retina beim menschen mittels der Hermannschen gittertäuschung. ," *Pflüger's Archiv für die gesamte Physiologie des Menschen und der Tiere,* vol. 272(1), pp. 21-22, 1960.

[21] D. M. W. Powers, "Jetwhite color map. Mathworks – https://au.mathworks.com/matlabcentral/fileexchange/48419-jetwhite-colours-/content/jetwhite.m. ," 2016.

[22] L. Spillmann, "Receptive fields of visual neurons: The early years," *Perception,* vol. 43, no. 11, pp. 1145-1176, 2014.

[23] F. Zöllner, "Über eine neue Art anorthoskopischer Zerrbilder," *Annalen der Physik,* vol. 193(11), pp. 477-484, 1862.

[24] T. Oyama, "Determinants of the Zöllner illusion," *Psychological research, 37(3), 261-280,* 1975.

[25] P. Bressan, "Revisitation of the family tie between Munsterberg and Taylor-Woodhouse illusions," *Perception, 14(5), 579-585,* 1985.

[26] M. E. McCourt, "Brightness Induction and the Cafe Wall Illusion," *Perception, 12(2), 131-142,* 1983.

[27] G. Westheimer, "Irradiation, border location, and the shifted-chessboard pattern," *Perception,* vol. 36, no. 4, pp. 483-94, 2007.

[28] A. Kitaoka, B. Pinna, and G. Brelstaff, "New variations of the spiral illusion," *Perception,* vol. 30(5), pp. 637-646, 2001.

[29] A. Kitaoka, "A Bulge," *http://www.ritsumei.ac.jp/~akitaoka/index-e.html,* 1998.

[30] N. Nematzadeh, T. W. Lewis, and D. M. W. Powers, "Bioplausible multiscale filtering in retinal to cortical processing as a model of computer vision," *ICAART2015-International Conference on Agents and Artificial Intelligence. SCITEPRESS,* 2015.

[31] M. Carandini, "Receptive Fields and Suppressive Fields in the Early Visual System," (in English), *Cognitive Neurosciences Iii, Third Edition,* vol. 313, pp. 313-326, 2004.

[32] D. Marr and S. Ullman, "Directional Selectivity and Its Use in Early Visual Processing," (in English), *Proceedings of the Royal Society Series B-Biological Sciences,* vol. 211, no. 1183, pp. 151-+, 1981.

[33] N. Nematzadeh and D. M. W. Powers, "A bioplausible model for explaining Café Wall illusion: foveal vs peripheral resolution," *In International Symposium on Visual Computing. Springer International Publishing.,* Conference - ISVC2016 -12th International Symposium on Visual Computing pp. 426-438, 2016.

[34] J. Illingworth and J. Kittler, "A Survey of the Hough Transform," (in English), *Computer Vision Graphics and Image Processing,* vol. 44, no. 1, pp. 87-116, Oct 1988.

[35] D. H. Hubel and T. N. Wiesel, "Receptive fields, binocular interaction and functional architecture in the cat's visual cortex," *J Physiol,* vol. 160, pp. 106-54, Jan 1962.

[36] H. B. Barlow, "Pattern recognition and the responses of sensory neurons," *Ann N Y Acad Sci,* vol. 156, no. 2, pp. 872-81, Apr 21 1969.

[37] F. W. Campbell and J. G. Robson, " Application of Fourier analysis to the visibility of gratings " *The Journal of physiology, 197(3), 551-566,* 1968.

[38] D. Marr, "Early processing of visual information," *Philos Trans R Soc Lond B Biol Sci,* vol. 275, no. 942, pp. 483-519, Oct 19 1976.

[39] D. Marr, S. Ullman, and T. Poggio, "Bandpass channels, zero-crossings, and early visual information processing," *J Opt Soc Am,* vol. 69, no. 6, pp. 914-6, Jun 1979.

[40] B. Blakeslee and M. E. McCourt, "A multiscale spatial filtering account of the White effect, simultaneous brightness contrast and grating induction," *Vision Res,* vol. 39, no. 26, pp. 4361-77, Oct 1999.

[41] K. Ghosh, S. Sarkar, and K. Bhaumik, "Early vision and image processing: evidences favouring a dynamic receptive field model," *In Computer Vision, Graphics and Image Processing. Springer Berlin Heidelberg,* pp. 216-227, 2006.

[42] Y. Q. Lv, G. Y. Jiang, M. Yu, H. Y. Xu, F. Shao, and S. S. Liu, "Difference of Gaussian Statistical Features Based Blind Image Quality Assessment: A Deep Learning Approach," (in English), *2015 Ieee International Conference on Image Processing (Icip),* pp. 2344-2348, 2015.